\pdfoutput=1
\documentclass[10pt, a4paper]{article}

\usepackage[final]{lrec-coling2024} % this is the new style

\usepackage{microtype}
\usepackage{times}
\usepackage{latexsym}
\usepackage{graphicx}  %Required
\usepackage{subfigure}
\usepackage{url}
\usepackage{amsfonts} 
\usepackage{amsmath}
\usepackage{booktabs}
\usepackage{multirow}
\usepackage{arydshln}
\usepackage{booktabs}
\usepackage{pgfplots}
\usepackage{subfigure}
\usepackage{hyperref}
\usepackage[T1]{fontenc}
\usepackage[utf8]{inputenc}

\usepackage{microtype}

\usepackage{inconsolata}

\title{Mitigating Misleading Chain-of-Thought Reasoning \\with Selective Filtering}

\name{Yexin Wu$^{1}$, Zhuosheng Zhang$^{2,*}$, Hai Zhao$^{3,4,*}$\thanks{$^*$Corresponding authors. This paper was partially supported by Joint Research Project of Yangtze River Delta Science and Technology Innovation Community (No. 2022CSJGG1400) and Joint Funds of the National Natural Science Foundation of China (No. U21B2020). }}
\address{$^1$College of Zhiyuan, Shanghai Jiao Tong University\\$^2$School of Electronic Information and Electrical Engineering, Shanghai Jiao Tong University\\ $^3$Department of Computer Science and Engineering, Shanghai Jiao Tong University\\ $^4$Key Laboratory of Shanghai Education Commission for Intelligent Interaction\\ and Cognitive Engineering, Shanghai Jiao Tong University\\ \texttt{\{wuyexin\_libro\_i131, zhangzs\}@sjtu.edu.cn, zhaohai@cs.sjtu.edu.cn}}
\abstract{Large language models have manifested remarkable capabilities by leveraging chain-of-thought (CoT) reasoning techniques to solve intricate questions through step-by-step reasoning chains. Despite its success, the efficacy of such reasoning is inherently contingent upon the quality of CoT. However, flawless CoT reasoning cannot be guaranteed due to the presence of indecomposable questions and the potential for erroneous reasoning chains, particularly in the case of small-scale language models. To tackle this challenge, we propose a novel approach called the selective filtering reasoner (SelF-Reasoner) that assesses the entailment relationship between the question and the candidate reasoning chain. Then, we proceed with CoT reasoning when the reasoning chain demonstrates confidence; otherwise, we opt to predict the answer directly. SelF-Reasoner improves the fine-tuned T5 baseline consistently over the ScienceQA, ECQA, and LastLetter tasks. Code is available at \texttt{https://github.com/LibroWu/SelF-Reasoner}.}
\begin{document}
\maketitleabstract

%Large language models have manifested remarkable capabilities by leveraging chain-of-thought (CoT) reasoning techniques to solve intricate questions through step-by-step reasoning chains. Despite its success, the efficacy of such reasoning is inherently contingent upon the quality of CoT. However, flawless CoT reasoning cannot be guaranteed due to the presence of indecomposable questions and the potential for erroneous reasoning chains, particularly in the case of small-scale language models. To tackle this challenge, we propose a novel approach called the selective filtering reasoner (SelF-Reasoner) that assesses the entailment relationship between the question and the candidate reasoning chain. We proceed with CoT reasoning when the reasoning chain demonstrates confidence; otherwise opting to predict the answer directly. SelF-Reasoner improves the fine-tuned T5 baseline by +13.1\% improvement (74.1\%$\rightarrow$87.2\%) over the ScienceQA benchmark. Code is available at \texttt{Anonymous}.

\section{Introduction}
\begin{center}
	\begin{tabular}{p{0.9\linewidth}}
		\noindent\emph{``I will select their good qualities and follow them, their bad qualities and avoid them.''} \\
		\midrule
		\hfill{\emph{ Confucius {\rm(551 BC - 479 BC)}}}\\
	\end{tabular}
\end{center}

Large language models \citep[LLMs]{brown2020language, lamda, gopher, palm} have exhibited impressive capabilities in various reasoning tasks, including arithmetic and symbolic reasoning, by generating intermediate chain-of-thought (CoT) reasoning steps \citep{nye2022show,cot_wei,kojima2022large}. 

Although CoT approaches have shown improvements in reasoning performance and interoperability, there are still two main challenges that hinder the widespread adoption of those approaches: (i) indecomposable questions, which refer to simple questions that cannot be decomposed into smaller sub-questions; (ii) erroneous reasoning chains, which involve mistakes in the logical and commonsense reasoning processes \citep{cot_wei,kojima2022large,auto-cot}, and may even result in hallucinations \citep{wang2022iteratively,zhang2023multimodal} or unfaithful explanations \citep{turpin2023language}.
\begin{figure}[t]
\centering
\includegraphics[width=\linewidth]{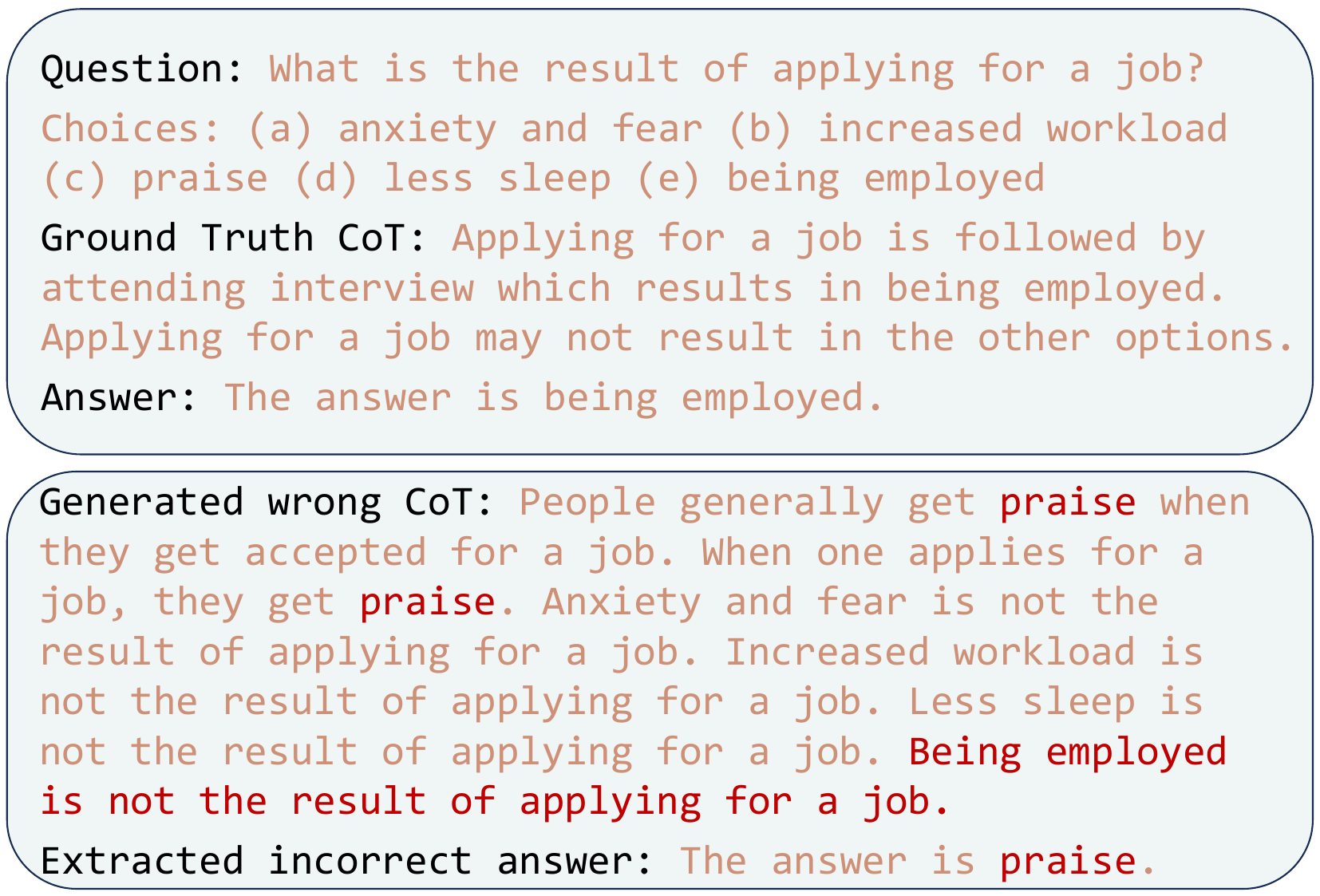}
\vspace{-0.15in}
\caption{An example of an invalid CoT reasoning from ECQA~\cite{aggarwal-etal-2021-explanations, wang2023pinto}. The errors are highlighted in \textcolor{red}{red}. The generated CoT is wrong at the first step, and the error continues to the end. However, when altering to direct prediction, this one-step reasoning question is solved correctly.}
\vspace{-0.15in}
\label{fig:example}
\end{figure}

%\begin{figure}[t]
%\centering
%\includegraphics[width=\linewidth]{figures/demo_example.pdf}
%\caption{An example of CoT reasoning from ScienceQA \citep{lu2022learn}.}
%\vspace{-0.15in}
%\label{fig:example}
%\end{figure}

The above imperfect nature of CoT essentially leads to inferior results when using CoT instead of directly predicting answers, especially for small-scale language models prone to generating flawed reasoning chains \citep{zhang2023multimodal}. Consequently, small-scale language models are unlikely to achieve better reasoning performance with self-generated CoT alone, unless distilling the effective CoT from LLMs to minimize the occurrence of erroneous reasoning chains \citep{Small_Reasoners_Better, teaching, teacher}.

Existing studies have proposed verification methods to improve the correctness of reasoning chains \citep{gsm8k,zhao2023verify,diao2023active,sun2023enhancing,lightman2023let}. However, they deal with all the questions concerned by CoT in general, without selectively discarding irrelevant CoTs when necessary. Besides, their primary focus lies in refining the reasoning chains, neglecting the entailment relationship between the question and candidate reasoning chain. To bridge the gap, this work aims to tackle both challenges as described above simultaneously.

In this work, we propose a novel approach called the selective filtering reasoner (SelF-Reasoner) that assesses the entailment relationship between the question and the candidate reasoning chain. SelF-Reasoner proceeds with CoT reasoning when the reasoning chain demonstrates confidence; otherwise opting to predict the answer directly. Figure~\ref{fig:example} shows a case when the generated CoT is wrong but the direct prediction can be correct. The SelF-Reasoner is composed of three key components: (i) a reasoner responsible for generating the candidate chain; (ii) an answerer module capable of predicting the final answer directly or extracting it from the question-solution pair; (iii) a CoT Filter designed to discard invalid reasoning chains and enhance the model's performance by utilizing the effective reasoning chain. 

We implement SelF-Reasoner on the fine-tuned T5 models, revealing that small-scale language models can also benefit from CoT if equipped with our selective filtering mechanism. Experimental results on benchmarks show that our proposed method SelF-Reasoner improves the fine-tuned T5 baseline consistently over the ScienceQA\citep{lu2022learn}, ECQA, and LastLetter tasks. In summary, our main contributions are as follows: %outperforms the previous fine-tuned T5 baseline by +13.1\% accuracy (74.1\% \cite{lu2022learn}$\rightarrow$87.2\%). In summary, our main contributions are as follows:

(i) We proposed a selective filtering reasoner (SelF-Reasoner) to perform CoT only as necessary and mitigate the detrimental effects of erroneous reasoning chains.    

(ii) Our SelF-Reasoner outperforms the fine-tuned CoT/vanilla baseline on ScienceQA, ECQA, and LastLetter datasets, advancing the effectiveness of CoT in small-scale language models.

(iii) We analyze the obstructions of fine-tuning CoT on language models and conclude common types in invalid generated CoT.

\begin{figure*}
\centering
\includegraphics[width=1.0\linewidth]{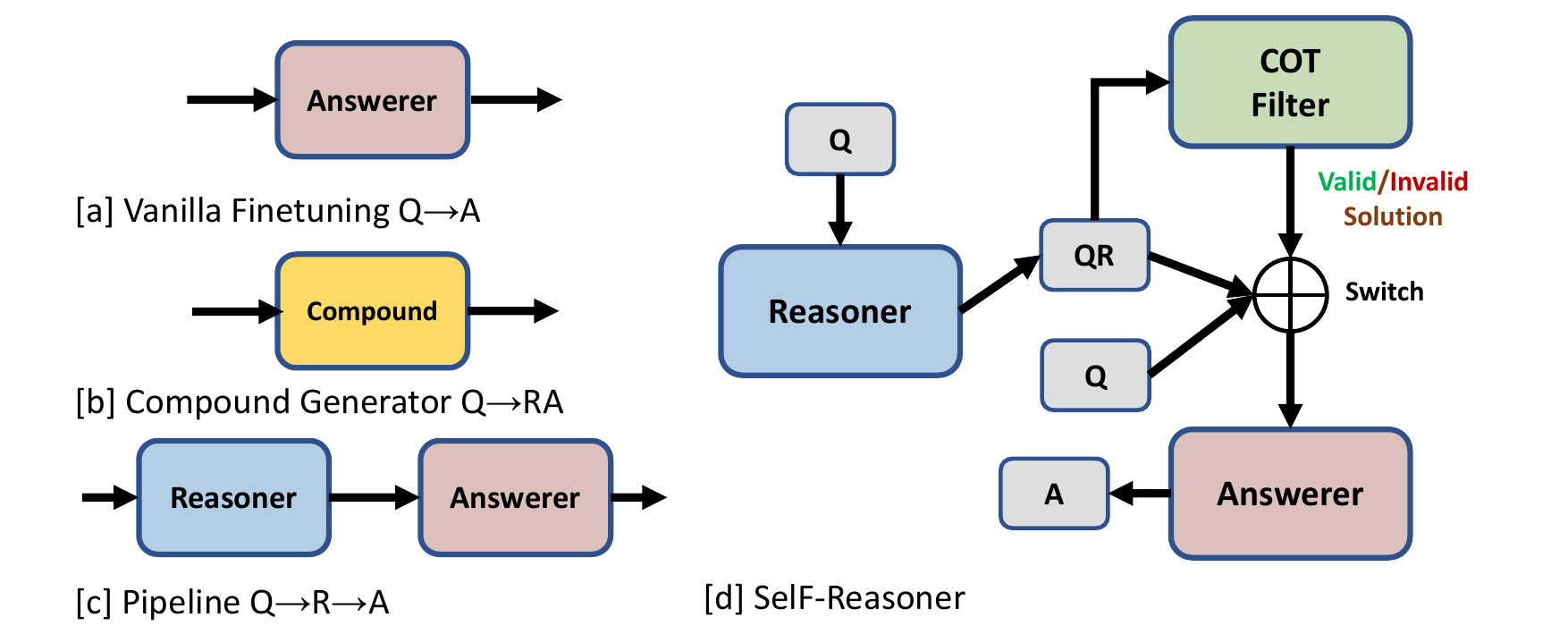}
\caption{(a) Vanilla fine-tuning model that directly predicts the answer. (b) A compound generator that produces the reasoning chain and the answer simultaneously. (c) The two-level pipeline produces the reasoning chain first and then extracts the answer from it. (d) \textbf{Our SelF-Reasoner}. The Reasoner is fine-tuned to generate a solution based on the input question description. The Answerer is the model that can predict the answer directly or extract the answer from a question-solution pair. CoT Filter can determine whether the solution generated can be passed to the Answerer. Here, Q stands for Question, R stands for Reasoning Chain, and A stands for Answer. The symbol QR means concatenating Q and R, RA means concatenating R and A.}
\label{fig:arch}
\end{figure*}

\section{Related Work}
Two lines of research are key to our work: chain of thought prompting and fine-tuning language models to reason.
\subsection{Chain-of-thought Prompting}
CoT prompting is a gradient-free strategy that encourages LLMs to generate the incremental thought processes required to reach a desirable outcome \citep{nye2022show,cot_wei,kojima2022large}. This approach aids LLMs in producing a chain of intermediate steps that leads to the final response to a question. According to the number of CoT examples in the prompt, we categorize CoT prompting into two primary methods: Zero-Shot CoT \citep{kojima2022large} and Few-Shot CoT \citep{cot_wei}.
\paragraph{Zero-Shot CoT.} \citet{kojima2022large} showed that LLMs are good zero-shot reasoners who already obtain the capability to reason from large-scale pretraining. A task-agnostic \textit{magic spell} like \textit{Let's think step by step} can make the LLMs generate the immediate reasoning steps and give the final answer. However, we do not find similar behavior in the smaller language models, mainly due to the limited training data and memory ability. Fine-tuning with a specific generation format further reduces the possibility of doing Zero-Shot CoT. So we fine-tune language models to generate reasoning chains instead of Zero-Shot CoT.
\paragraph{Few-Shot CoT.} Few-Shot CoT achieves more robust performance by harnessing reasoning ability with the help of specially designed or self-created reasoning demonstrations. Researchers have been looking into various strategies to enhance the effectiveness of Few-Shot CoT. Least-to-most prompting \citep{zhou2022least} decomposes complex problems into more manageable sub-problems to be solved sequentially. \citet{cot_wei_sc} introduced a self-consistency decoding strategy in that various reasoning trajectories are sampled to vote for the final answer. \citet{gsm8k} further trained verifiers on math problems to rank the candidate reasoning chains to get the most reliable reasoning chain. However, significant effort is still necessary to build the human-annotated reasoning chains in these methods. To overcome this limit, \citet{auto-cot} proposed Auto-CoT utilizing Zero-Shot CoT and clustering to generate reasoning demonstrations automatically. 

The performance of Few-Shot CoT can be influenced by the context and quality of the thought processes involved in the demonstrations. On certain specific tasks, utilizing fine-tuned language models may result in superior performance compared to using few-shot prompting LLMs.

\subsection{Fine-tuning Language Models to Reason}
Inspired by the success of CoT in LLMs, we investigate the possibility of applying CoT fine-tuning to smaller language models. Concurrent works \citep{Small_Reasoners_Better, teaching, teacher, li2022explanations, wang2023pinto} mainly utilize LLMs to create reliable reasoning chains, but these techniques still depend on LLMs. \citet{hsieh2023distilling} frames CoT learning as multi-task learning with rationales and shows their method can reduce the model size and the training data to achieve better performance than LLM. \citet{wang2023pinto} utilizes LLM to generate CoT and fine-tunes a smaller answerer under counterfactual training to make sounder use of the rationales. However, these works don't discard the produced invalid CoT even if it will lead to an incorrect answer. In our work, we focus on the fine-tuning process, and, different from \citet{wang2023pinto}, we explicitly introduce a CoT filter to alleviate the effects of misleading CoTs. 

\citet{lu2022learn} constructed a multimodal benchmark called Science Question Answering (ScienceQA) and fine-tuned language models to produce reasoning chains utilizing the annotated rationales in the dataset. They found that smaller language models, such as UnifiedQA \citep{unifiedQA}, can benefit from CoT fine-tuning. However, in their setting, the model first provides the answer and then the reasoning chain, which only focuses on rationalization like WT5 \citep{wt5}. In this paper, we thoroughly examine the CoT fine-tuning on this dataset and reflect on the provided reasoning chains.

\section{Approach}\label{sec:approach}
This section will first present the details of our baseline for comparisons and our design of the SelF-Reasoner.

\subsection{Baselines}
We conduct experiments on three baselines: vanilla fine-tuning, compound generator, and two-level pipeline. For simplicity, let \textit{Q} denote \textit{Question}, \textit{R} denote \textit{Reasoning Chain}, and \textit{A} denote \textit{Answer}. Format \textit{Q-R} means to input the question and generate the reasoning chain.

\paragraph{Vanilla Fine-tuning.} In this setting (Figure~\ref{fig:arch} (a)), we fine-tune the model only to generate the final answer (format \textit{Q-A}). Concretely, given the input question, the model generates the answer directly: 
\begin{equation}
    A = \textbf{Answerer}(Q).
\end{equation}

\paragraph{Compound Generator.} \label{sec:def} The compound generator (Figure~\ref{fig:arch} (b)) produces the reasoning chain and final answer simultaneously. Given the input question, the compound generator generates an output sequence with the reasoning chain and answer:
\begin{equation}
    \{R,A\} = \textbf{Compound}(Q).
\end{equation}

Depending on the positions of the reasoning chain and answer, we name the format where the reasoning chain is before the answer (\textit{Q-RA}) as the CoT fine-tuning and the format where the reasoning chain is after the answer (\textit{Q-AR}) as the reasoning chain fine-tuning. We will show that the practice of compound CoT fine-tuning is beset with the possibility of missing answers in the output (Section~\ref{sec:results}). To rectify this problem, we propose the implementation of a two-level pipeline, which will be presented subsequently.

\paragraph{Two-level Pipeline.} In the two-level pipeline (Figure~\ref{fig:arch} (c)), there are a reasoner and an answer extractor. The reasoner produces only the reasoning chain and appends it to the question. Then the answer extractor extracts the final answer:
\begin{eqnarray}
    R & = & \textbf{Compound}(Q), \\
    A & = & \textbf{Answerer}(Q \circ R). 
\end{eqnarray}

The reasoner is fine-tuned in the format \textit{Q-R}, and the extractor is fine-tuned in the format \textit{QR-A}. We will show that the reasoner can provide misleading reasoning chains, leading to the answer extractor producing incorrect responses (Section~\ref{sec:results} \& \ref{sec:CoT}). In light of this finding, a CoT filter that identifies invalid CoTs has been integrated into the system to form the SelF-Reasoner introduced in the following subsection.
\subsection{SelF-Reasoner}
The SelF-Reasoner (Figure~\ref{fig:arch} (d)) consists of a reasoner, an answerer, and a CoT filter. The reasoner is fine-tuned on the format \textit{Q-R}:
\begin{equation}
    R = \textbf{Compound}(Q).
\end{equation}

The CoT filter will determine whether the reasoning chain is valid, thus filtering out incorrect reasoning chains. To train the CoT verifier, we use the generated CoT from the two-level pipeline to construct \textit{QR-label} pair. If the generated CoT leads to the correct answer, then label this CoT as valid (label=1). Otherwise, the label is invalid (label=0):
\begin{equation}
    v = \textbf{Verify}(Q \circ R).
\end{equation}

The answerer is fine-tuned on the format \textit{Q-A} augmented with the format \textit{QR-A}, so it can directly predict the final answer or extract the answer from the question-CoT pair. 
\begin{equation}
A =  \begin{cases}
\textbf{Answerer}(Q \circ R),& v =1 \\ 
\textbf{Answerer}(Q),& v = 0.
\end{cases}
\end{equation}%

To train the verifier, we use the T5 \citep{t5} encoder extended with a linear classification layer as our verifier model. We load the convergent checkpoint of the answerer as the initial parameters before fine-tuning if possible. It can accelerate the training procedure and improve convergent accuracy.

%To construct the training data, we run the pipeline model on the training data. The instances with reasoning chain leading to correct predictions will be labeled as positive and those leading to incorrect predictions will be labeled as negative. To balance the number of positive and negative instances, we use a subpar reasoner which scores less precise in the training set to generate invalid instances.\footnote{In ScienceQA, the convergent two-level pipeline can reach a score of 90\% in the training set. So we use a subpar reasoner which scores 60\% in the training set to generate invalid instances.}
We include more implementation details in Appendix~\ref{Appendix: Verifier Training Details}.

The verifier can also be logical rules to avoid extra computation. We use the rule that \textbf{"the given word should appear in the valid CoT"} to filter out invalid CoT in the LastLetter Task.
%We load the convergent checkpoint of the answerer before fine-tuning if the filter and the answerer use the same model. If the backbone model architecture is T5 \citep{t5}, we add a linear classification layer to the T5 encoder. %If the backbone model is BERT \citep{bert}, we add a linear classification layer to the BERT model.
\section{Experiments}
This section will describe our experimental setup regarding datasets, language models, and evaluation, and present the main results.
\subsection{Datasets}
We use ScienceQA~\citep{lu2022learn}, ECQA~\citep{aggarwal-etal-2021-explanations, talmor-etal-2019-commonsenseqa} and LastLetter for our experiments. ScienceQA is a benchmark that consists of multimodal multiple-choice questions with a diverse set of science topics and annotations of their answers with corresponding lectures and explanations.  ECQA is a human-annotated version of CommonsenseQA~\citep{talmor-etal-2019-commonsenseqa}, which is a 5-choice QA dataset assessing general commonsense reasoning using concepts from ConceptNet \citep{speer2018conceptnet}. LastLetter is the task of concatenating the last letters of given words. More details about datasets can be found in Appendix~\ref{Appendix: Dataset}.

%ScienceQA consists of 12k training problems and 4k evaluation/test problems. ECQA annotates CommonsenseQA~\citep{talmor-etal-2019-commonsenseqa} by human experts. It consists of 8.5k training problems and 1.2k evaluation/test problems. 

%We also train baselines on two arithmetic reasoning datasets, AQuA~\citep{aqua} and GSM8K~\citep{gsm8k}. All two datasets have human-annotated reasoning chains for each data example.
\subsection{Language Models}
We use UnifiedQA-T5 \citep{unifiedQA} as our primary language model to align with the experiments on the ScienceQA paper. Also, the imperfect nature of CoT is more salient in such language models. We select UnifiedQA-T5-small (60M), UnifiedQA-T5-base (220M), and UnifiedQA-T5-large (770M). More implementation details are included in Appendix~\ref{Appendix: Implementation Details}.
\subsection{Evaluation}
We use the accuracy of the answer (exactly matching for produced answer sentence) to measure the performance. To measure the quality of the generated reasoning chain, we use automatic metrics, such as BLEU-1/4 \citep{bleu}, ROUGE-L~\citep{bleu}, and sentence similarity, as the settings on the ScienceQA paper. We use the Sentence-BERT network~\citep{reimers-2019-sentence-bert} to encode the generated and ground truth reasoning chains and compute the cosine similarity as the metric Similarity. 

Besides the automatic evaluations, we further conduct a human study to investigate our model performance more comprehensively. We randomly sample 50 instances from the test set and evaluate the entailment, completeness, and correctness. Entailment means the CoT entails the extracted answer. Completeness means that the CoT is complete. Correctness means that the CoT is correct and relevant to the question.

\subsection{Main Results}\label{sec:results}
In this section, we will present the main results and observations of our baselines, followed by a discussion of our SelF-Reasoner. Table~\ref{tab:accuracy} presents the main results of SelF-Reasoner compared with our baselines and existing methods of the ScienceQA benchmark. Tabel~\ref{tab:ecqa} presents the results of the ECQA dataset. Table~\ref{tab:lastletter} presents the results of the LastLetter task.

\begin{table*}[h]
    %Table for accuracy, ScienceQA
    \centering
    % \resizebox{\textwidth}{!}
    \small
    \setlength{\tabcolsep}{10pt}
    {
    \begin{tabular}{llccc}
      \toprule
         Method & Model & Learning & Format & Accuracy\\
         \midrule
        \multirow{5}{6em}{\citet{lu2022learn}} & Human & - & \textit{Q-A}& 88.4\\
        & GPT-3 (CoT)& In-Context Learning& \textit{Q-ALE}& 75.17\\
        & UnifiedQA$_\textup{Base}$& fine-tuning & \textit{Q-A}& 70.12\\
        & UnifiedQA$_\textup{Base}$& fine-tuning & \textit{Q-AE}& 73.33\\
        & UnifiedQA$_\textup{Base}$& fine-tuning & \textit{Q-ALE}& 74.11\\
        \midrule
        \multirow{4}{6em}{\citet{lu2023chameleon}}& ChatGPT (CoT)& In-Context Learning & \textit{Q-EA}& 78.31\\
        & GPT-4 (CoT)& In-Context Learning & \textit{Q-EA}& 83.99\\
        & Chameleon (ChatGPT)& In-Context Learning & \textit{Q-EA}& 79.93\\
        & Chameleon (GPT-4) & In-Context Learning & \textit{Q-EA}& 86.54\\
        \midrule
        \multirow{3}{4em}{Vanilla} & UnifiedQA$_\textup{Samll}$& fine-tuning & \textit{Q-A}& 71.54\\
        & UnifiedQA$_\textup{Base}$& fine-tuning & \textit{Q-A}& 83.09\\
        & UnifiedQA$_\textup{Large}$& fine-tuning & \textit{Q-A}& 86.53\\
        \midrule
        \multirow{3}{4em}{Compound} & UnifiedQA$_\textup{Base}$& fine-tuning & \textit{Q-ALE}& 76.13\\
        & UnifiedQA$_\textup{Base}$& fine-tuning & \textit{Q-EA}& 77.71\\
        & UnifiedQA$_\textup{Base}$& fine-tuning & \textit{Q-LEA}&  73.97 \\
        \midrule
        \multirow{3}{4em}{Pipeline} & UnifiedQA$_\textup{Small}$& fine-tuning & \textit{Q-E}$\rightarrow$ \textit{QE-A}& 66.37\\
        & UnifiedQA$_\textup{Base}$& fine-tuning & \textit{Q-E}$\rightarrow$ \textit{QE-A}& 79.32\\
        & UnifiedQA$_\textup{Large}$& fine-tuning & \textit{Q-E}$\rightarrow$ \textit{QE-A}& 84.98\\
        \midrule
        \multirow{3}{8em}{SelF-Reasoner} & UnifiedQA$_\textup{Small}$& fine-tuning & SelF-Reasoner & 69.55\\
        & UnifiedQA$_\textup{Base}$& fine-tuning & SelF-Reasoner & 83.45\\
        & UnifiedQA$_\textup{Large}$& fine-tuning & SelF-Reasoner & \textbf{87.24}\\
\bottomrule
    \end{tabular}
    }
    \caption{ Accuracy (\%) of each baseline on test split. In the format part, \textit{Q = Question}, \textit{A = Answer}, \textit{E = Explanation}, \textit{L = Lecture}. We list the results from ScienceQA \citep{lu2022learn}, ChatGPT, GPT-4 \citep{lu2023chameleon} for comparison. \textit{L} and \textit{E} can be treated as reasoning chain. So LEA/EA and ALE/AE correspond to the standard RA and AR as defined in Section~\ref{sec:def}, respectively. Our SelF-Reasoner (Large) is comparable in accuracy to a human's.}
    \label{tab:accuracy}
    \vspace{-0.05in}
\end{table*}
% todo: AQuA and GSM8k

\paragraph{Vanilla Fine-tuning is a Strong Baseline.} On ScienceQA benchmark, UnifiedQA$_\textup{Large}$ achieves an accuracy of 86.53\%, which is close to the human performance from Table~\ref{tab:accuracy}, indicating that vanilla fine-tuning is a strong baseline.\footnote{There is a 13\% discrepancy between the performance of vanilla fine-tuning reported by ours and \citet{lu2022learn}. The discrepancy is discussed in Appendix~\ref{Appendix: Gap}.} Despite the high accuracy, vanilla fine-tuning still lacks interpretability. Then our next objective is to further elicit the model's reasoning capacity to explain its thought process while maintaining its accuracy.

% We further operate the experiments on the AQuA and GSM8k datasets. The detailed results of these two datasets can be found in Appendix~\ref{sec:baseline}. % todo: brief result introduction

\paragraph{Compound Generator Suffers from Imperfect CoT.} The accuracy of CoT fine-tuning is lower than that of vanilla fine-tuning, in contrast to the benefits of CoT prompting in LLMs. The main reason is that the loss of the answer part is weakened by the CoT part. We also observe that the outputs are too lengthy in the setting of RA; thus, answers are not given due to the maximum length limits. We find that the ratios of missing answers in the training set are 0.8\% in format \textit{Q-EA} and 5.9\% in format \textit{Q-LEA}. Due to the answer missing issue, the performance of the \textit{Q-RA} model is inferior to that of the \textit{Q-AR} model, in agreement with the findings reported in \citet{lu2022learn}. However, we do not adopt the AR format because it is more of a posterior rationalization than a CoT.

\begin{table*}[t]
    % table for ScienceQA generated CoT metrics
    \centering
    % \resizebox{\textwidth}{!}
    \small
    \setlength{\tabcolsep}{4.2pt}
    {
    \begin{tabular}{cc|ccc|c|ccc}
      \toprule
         Model & Split & BLEU-1 & BLEU-4 & ROUGE-L & Similarity & Complete& Entailment & Correct \\
         \midrule
        \multirow{3}{4em}{Base} & Lead to Correct Answer &  0.914 & 0.776 & 0.910 & 0.937 & 1.00&1.00&0.94\\
         & Lead to Incorrect Answer &  0.789 & 0.660 & 0.797 &  0.860 & 1.00&1.00& 0.02\\
         & All & 0.892 & 0.756 & 0.891 & 0.924 & -&-&-\\
                  \midrule
        \multirow{3}{4em}{Large} & Lead to Correct Answer &  0.937 & 0.810 & 0.929 & 0.949 &1.00 &0.98&0.96\\
         & Lead to Incorrect Answer &  0.775 & 0.642 & 0.784 &  0.847 & 1.00&1.00& 0.02\\
         & All & 0.917 & 0.788 & 0.910 & 0.936 &- &-& -\\
\bottomrule
    \end{tabular}
    }
    \caption{ Automatic metrics (BLEU-1/4, ROUGE-L, Similarity) and human evaluation of generated explanations. We evaluate these metrics on different splits of the produced CoT according to whether they can lead to the correct answer. Details of human evaluation are shown in Appendix~\ref{Appendix: HumanEval}.
    }
    \label{tab:generated_evaluate}
    \vspace{-0.15in}
\end{table*}

%(iii) By assessing the effectiveness of lectures and solutions through fine-tuning the \textit{QCML-A} and \textit{QCME-A} models, considerable differences in accuracy were observed. The \textit{QCML-A} model yielded an accuracy of 79\%, whereas the \textit{QCME-A} reported an accuracy of 99\%. This demonstrates that the solution had greater significance than the lecture. As such, for further experiments, we shall solely rely on the solution, thus simplifying the training task.

%In order to avoid any missing answers in the training targets, we created two models (\textit{QCM-E} and \textit{QCME-A}) from the \textit{QCM-EA} format and connected them together in a pipeline that will be inspected in the following section.

% We further operate the experiments on the AQuA~\citep{aqua} and GSM8k~\citep{gsm8k} datasets. The detailed results of these two datasets can be found in Appendix~\ref{sec:baseline}.% todo: brief result introduction

\paragraph{The Pipeline Method Narrows the Gap in Performance.} The pipeline method narrows the gap between the compound generator and the vanilla fine-tuning (Table~\ref{tab:accuracy}). We conduct a thorough evaluation of the produced reasoning chain in Section~\ref{sec:CoT}. By evaluating the respective cases, we find that some questions that the pipeline fails to answer can be properly solved by the vanilla fine-tuning model and vice versa. Vanilla fine-tuning and CoT fine-tuning can cause the model to acquire different segments of knowledge. This property can be utilized to augment the system's performance.

%By utilizing this property, we put forward the selective-filtering reasoner (SelF-Reasoner) to enhance the pipeline's performance. We use an additional model, called the CoT Filter, to identify invalid CoTs and prevent them from distorting the result. The subsequent section will present the result of this method in more detail.

% We further operate the experiments on the AQuA and GSM8k datasets. The detailed results of these two datasets can be found in Appendix~\ref{sec:baseline}. % todo: brief result introduction

% refer to the analysis of the filter later.
\paragraph{SelF-Reasoner Performs the Best.} Our SelF-Reasoner gets the best performance consistently over the ScienceQA, ECQA, and LastLetter tasks. 

On the ScienceQA benchmark (Table~\ref{tab:accuracy}), the base and large SelF-Reasoner models guarantee significant improvement over the pipeline, and slightly outperform the strong vanilla fine-tuning model under both base and large sizes, respectively. Specifically, our best SelF-Reasoner model outperforms the previous state-of-the-art model (among the text-only models) Chameleon (GPT-4). Our SelF-Reasoner also significantly outperforms GPT-3 (CoT) by 12.07\% and ChatGPT (CoT) by 8.93\%, using a much smaller model size. %The average increment of accuracy among different model sizes is 3.19\%. 

SelF-Reasoner also performs the best on ECQA. Table~\ref{tab:ecqa} presents the accuracy performance on the ECQA dataset. Though the T5-base model can not learn to produce CoT as well as human experts, the verifier alleviates the insufficiency and boosts the accuracy. SelF-Reasoner successfully meets our goal of creating a highly accurate and interpretable model. There is still a gap between the SelF-Reasoner and PINTO (61.67\% as reported in \citet{wang2023pinto}) because PINTO gets LLMs involved to produce the rationale at test time, which can cause large computation or memory costs. In contrast, our approach does not rely on LLMs; thus it is more generally effective.

On the LastLetter task (Table~\ref{tab:lastletter}), where the pipeline outperforms the vanilla fine-tuning, SelF-Reasoner can still benefit the performance. We apply the simple rule that \textbf{"the given word should appear in the valid CoT"} to filter out invalid CoTs because we find that the reasoner often replaces the given word due to randomness in sampling. For example, "speakers" can be replaced by "speaking". Without much overhead, the SelF-Reasoner outperforms the pipeline by 3.26\%.

Overall, the results show that SelF-Reasoner successfully meets our goal of creating a highly accurate and interpretable model. The positive outcome indicates the significance of the CoT filter in fine-tuning language models. A detailed analysis of the pipeline component and the CoT filter has been conducted in Section~\ref{sec:CoT} and Section~\ref{sec:filter}, respectively.

\begin{table}[t]
    \centering
    \small
    \setlength{\tabcolsep}{2pt} 
    {
    \centering
        % \resizebox{\columnwidth}{!}
        {
    \begin{tabular}{lccc}
    \toprule
      Method &Vanilla & Pipeline &  SelF-Reasoner  \\% & sum\\
     \midrule
      Accuracy & 58.07 & 54.95 & \textbf{58.48 (+3.5)}\\
         \bottomrule
    \end{tabular}}}
    \caption{ Accuracy (\%) on test split of ECQA. The backbone model is UnifiedQA-base. SelF-Reasoner outperforms the pipeline by 3.5\%.}
    \label{tab:ecqa}
    \vspace{-0.15in}
\end{table}

\begin{table}[t]
    \centering
    \small
    \setlength{\tabcolsep}{2pt} 
    {
    \centering
        % \resizebox{\columnwidth}{!}
        {
    \begin{tabular}{lccc}
    \toprule
      Method &Vanilla & Pipeline &  SelF-Reasoner  \\% & sum\\
     \midrule
      Accuracy & 64.22 & 76.80 & \textbf{80.06 (+3.26)}\\
         \bottomrule
    \end{tabular}}}
    \caption{ Accuracy (\%) on test split of LastLetter. The backbone model is UnifiedQA-base. SelF-Reasoner outperforms the pipeline by 3.26\%.}
    \label{tab:lastletter}
    \vspace{-0.15in}
\end{table}

\section{Analysis}
To understand how SelF-Reasoner works and gain insights, we analyze the generated reasoning chains and the influence of the CoT filter. An incorrect CoT generated and filtered out by SelF-Reasoner is shown in Figure~\ref{fig:example}. More cases can be found in Appendix~\ref{Appendix:case}, Table~\ref{tab:ScienceQA}, Table~\ref{tab:ECQA}, and Table~\ref{tab:demo_lastletter} in Appendix.
%\subsection{Generality on Other Reasoning Tasks}
%To verify the general effectiveness of our method across tasks, we apply our methods in two datasets of arithmetic reasoning, AQuA and GSM8K. The results in Table \ref{tab:aqua_gsm8k} verify that our filtering method also improves model performance for the arithmetic reasoning.
%\begin{table}[htb]
%    \centering
    % AQuA & GSM8k accuracy
    % \vspace{-5mm}
    % \setlength{\tabcolsep}{20pt}
    % \resizebox{\textwidth}{!}%
%    \small
%    \setlength{\tabcolsep}{4pt} 
%    {
%    \centering
        % \resizebox{\columnwidth}{!}
%        {
%    \begin{tabular}{lccc}
%    \toprule
%      Dataset & Model & Format  &  Accuracy \\
%     \midrule
%      \multirow{4}{4em}{GSM8k} & \multirow{4}{4em}{Base} & Q-A & 5.00 \\
%      & & Q-RA & 5.68 \\
%      & & Q-AR & 4.32\\
%      & & Pipeline & 5.76\\
%     \midrule 
%      \multirow{5}{4em}{AQuA} & \multirow{4}{4em}{Base} & Q-A & 36.20 \\
%      & & Q-RA & 20.80 \\
%      & & Q-AR & 26.30\\
%      & & Pipeline & 32.28\\
%      & \multirow{1}{4em}{Large} & Pipeline & 36.61 \\
%         \bottomrule
%    \end{tabular}}}
%    \caption{ Accuracy(\%) of each baseline on test split.}
%    \label{tab:aqua_gsm8k}
%\end{table}
\subsection{Analysis on Generated Reasoning Chains}
\label{sec:CoT}

% todo: Case analysis
% table context, the difference between lead to correct and incorrect answers. refer to AQuA and GSM8k result in the appendix
% Relevant, correct, complete.
% analysis on incorrect CoTs, type of errors: caption wrong, the cot is not good, and so on. the obstructions part.
In this section, we evaluate the pipeline component of SelF-Reasoner by analyzing the produced reasoning chains. The evaluation includes both automatic metrics and human evaluation, and 50 examples from each data split are sampled for the human evaluation. The results show that the system makes some typical mistakes in the primary parts of the reasoning chains it produces. We also discuss the obstruction to generating perfect reasoning chains.
%(we present a list of our generated reasoning chains in Appendix~\ref{sec:CoT_list})
\paragraph{CoT fine-tuned model can produce invalid reasoning chains.}
In Table~\ref{tab:generated_evaluate}, BLEU and ROUGE metrics of correct samples are higher than the ones of incorrect samples, suggesting a quality gap in produced reasoning chains. The high Similarity metric (over 0.8) of invalid CoT indicates that the structure of the invalid CoT is similar to the ground truth. The human evaluation draws the same conclusion. 

The case study illustrates that the key and necessary objects, described as "bridging objects" in \citet{towards}, of the invalid reasoning chain mainly differ from the ground truth. We conclude some typical mistakes: (i) The bridging objects are missing; (ii) The bridging objects are mismatched; for example, feature A is attributed to object B, and feature B is attributed to object A; (iii) The bridging objects are wrong. These mistakes suggest the small language models are deficient in memorizing knowledge.

\paragraph{Incorrect reasoning chains can be traced back to both the model itself and the training data used.}
%The reason the CoT model makes mistakes can be attributed to the model and the training data. 

The model struggles to memorize all the necessary knowledge for the reasoning process, leading to errors in crucial parts of the reasoning chain. 

% todo: obstructions in the model. 
Our analysis of the training data revealed that certain reasoning chains on the ScienceQA dataset do not precisely conform to the chain-of-thought format. More discussion can be found in Appendix~\ref{Appendix: CoTFormat}. Potential future work could focus on the impact of the reasoning chain's format on the CoT fine-tuning.

\subsection{Influence of the CoT Filter}
\label{sec:filter}

\begin{table}[t]
    \centering
    % filter
    % \vspace{-5mm}
    % \setlength{\tabcolsep}{20pt}
    % \resizebox{\textwidth}{!}%
    \small
    \setlength{\tabcolsep}{6pt} 
    {
    \centering
        % \resizebox{\columnwidth}{!}
        {
    \begin{tabular}{lcccc}
    \toprule
      Model & Vanilla &  Pipeline &  Random& SelF-Reasoner\\
     \midrule
        Small & 71.54 &66.37  & 68.76 & 69.55\\
        Base &  83.09& 79.32 & 81.61 & 83.45\\
        Large & 86.53& 84.98 & 86.09 &87.24\\
         \bottomrule
    \end{tabular}}}
    \caption{Ablation on the CoT filter on ScienceQA benchmark. Random refers to randomly choosing vanilla fine-tuning and pipeline to produce the answer.}
    \label{tab:random}
    \vspace{-0.15in}
\end{table}
\begin{table}[t]
    \centering
    % verifier accuracy
    % \vspace{-5mm}
    % \setlength{\tabcolsep}{20pt}
    % \resizebox{\textwidth}{!}%
    \small
    \setlength{\tabcolsep}{3pt} 
    {
    \centering
        % \resizebox{\columnwidth}{!}
        {
    \begin{tabular}{lccccc}
    \toprule
      Generator & Filter & Valid Acc &  Invalid Acc  & Acc & F1\\
     \midrule
      \multirow{2}{4em}{Base} & Base & 76.96& 76.39 & 76.84 & 0.841\\
       & Large & 81.30 & 81.64 & 81.37 & 0.874 \\
     \midrule 
    \multirow{2}{4em}{Large} & Base & 74.97& 75.03 & 74.98& 0.836\\
       & Large & 80.07 & 78.17 & 79.78 & 0.871 \\
         \bottomrule
    \end{tabular}}}
    \caption{Accuracy and F1 score of the CoT filter on classifying the generated reasoning chain on ScienceQA benchmark. Valid/Invalid Acc refers to the filter's accuracy in discriminating valid/invalid reasoning chains. Acc is the overall accuracy.}
    \label{tab:verifier_accurac}
    \vspace{-0.15in}
\end{table}

% the impact from the big table(brief, avoid repeat the SelF-Reasoner part )
% mixing matrix, upper bound
% todo: Case analysis
%\subsection{Ablation Study of the CoT Filter}
This section discusses the CoT filter component of SelF-Reasoner. The influence of the filter on the pipeline is evaluated, and the upper bound capabilities of SelF-Reasoner with an ideal filter are discussed.
%This section discusses the CoT filter component of SelF-Reasoner. We evaluate the influence of the filter compared to the pipeline. Also, we discuss the upper bound of the SelF-Reasoner with an ideal filter.

\paragraph{The CoT Filter's Contribution to Better Accuracy} Table~\ref{tab:random} demonstrates the ablation on the CoT filter. SelF-Reasoner constantly outperforms the random baseline, indicating the significance of the filter. 

We further evaluate the prediction accuracy of the CoT filter for classifying the produced CoT from the base and large generators (Table~\ref{tab:verifier_accurac}). The filters demonstrate comparable capability in discriminating valid and invalid reasoning chains, though there is still potential for advancement. About 30\% CoTs are filtered out by the filter, while the invalid CoT rate is about 25\%.

\begin{figure}[t]
\centering
\includegraphics[width=\linewidth]{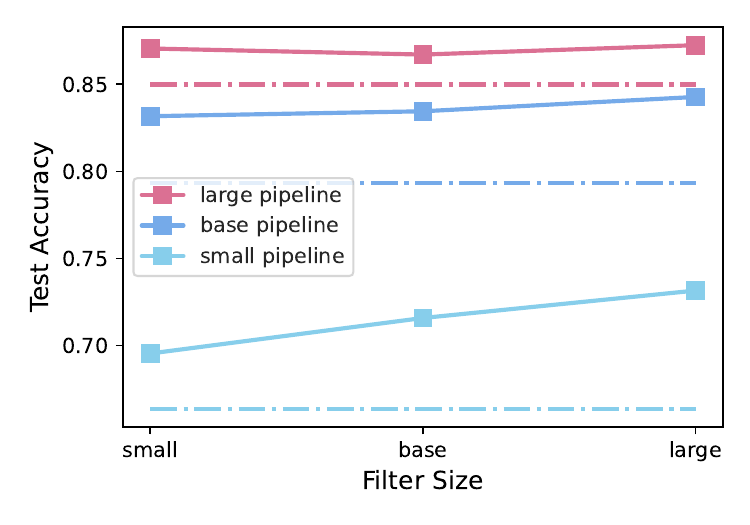}
\vspace{-0.40in}
\caption{The "scaling law" in the size of the CoT filter and pipeline on ScienceQA benchmark. The dashed line presents the accuracy of the pipeline.}
\label{fig:filter_size}
\vspace{-0.20in}
\end{figure}

\paragraph{"Scaling Law" of SelF-Reasoner} Figure~\ref{fig:filter_size} shows the "scaling law" in the size of the CoT filter and pipeline. A CoT filter can consistently enhance the accuracy of pipelines. As the filter is made larger, the improvement can be more significant; however, augmenting the size of a pipeline often leads to diminishing returns in terms of improvement, suggesting that a larger generator can produce invalid reasoning chains that are more similar to valid ones, thereby making it harder for the filter to differentiate them. 

%In addition, the confusion matrices in Table~\ref{tab:mixing} demonstrate the errors made by SelF-Reasoner. There is a certain amount of complex questions that can not be solved using both methods, indicating that the upper bound of SelF-Reasoner with an ideal filter is approximately 89\% for base and 91\% for large. 
We include the discussion on the upper bound of the Self-Reasoner in Appendix~\ref{Appendix: upperbound}.

% todo: CoT filter: T5 vs BERT. 

%\subsection{Out-of-Distribution Test}
% todo: OOD after the whole paper is done.
\section{Discussion}
In the discussion section, we delve into several key aspects related to CoT fine-tuning and the challenges associated with optimizing the performance of SelF-Reasoner.
\subsection{CoT Fine-tuning in a Specific Task} 
In this work, we observe that LLMs like GPT-3 may not consistently outperform fine-tuned small language models, even in tasks that involve generating complex and lengthy chain-of-thought sequences. This may be attributed to the fact that randomly chosen CoT prompts for LLMs may not effectively elicit optimal performance from LLMs, as the process of constructing such prompts can be both time-consuming and costly. Additionally, utilizing a small language model to achieve superior performance in a specific task is often a more straightforward approach regarding training and deployment.
\subsection{The Role of the Reasoning Chain} 
Through the case study, we find that the small language model benefits from templated reasoning chains due to its limited capacity. Despite the success of our model on the three datasets, the specific role of the reasoning chain within the prediction procedure remains uncertain. To further investigate the impact of the reasoning chain, future research may employ adversarial techniques such as incorporating noise and altering key parts ("bridging objects") or the reasoning templates, as outlined in \citet{towards}. This will aid in determining whether the improvement comes from the extra information introduced by the reasoning chain during the training phase or if the reasoning structure plays a significant role. Moreover, potential future work could focus on the impact of the reasoning chain's format on the CoT fine-tuning.

\subsection{The Role of CoT Filter}
The comprehensive evaluation of the reasoning chain is still dependent on human labor to examine individual sentences. Assessing the correctness of the reasoning chain, instead of merely determining if it leads to the correct answer, remains a complex task. The CoT filter enhances the accuracy and interpretability of the SelF-Reasoner, enabling the exclusion of misleading CoTs. However, the method of detecting invalid reasoning chains from valid ones is not currently understood. Future research can focus on developing interpretable filtering techniques utilizing rationalization methods and creating a toolkit for analyzing the quality of reasoning chains to aid in advancing CoT studies.

\subsection{Obstructions on the Way to Perfect CoT} Small language models may encounter significant obstacles in generating both reasoning chains and answers in a single turn, owing to the restricted maximum input and output length. Furthermore, it can be difficult for these models to maintain coherence in longer output sequences. Due to their limited number of parameters, small language models may suffer from difficulties in fully internalizing the intricate relationships present within the training data. This can result in a weaker understanding of the relationship between questions and answers when the primary focus is on learning the structure and complex relationships within reasoning chains. Additionally, it should be noted that not all tokens within a reasoning chain hold equal importance. This can lead to the model utilizing its already insufficient parameters to retain less important tokens and potentially overlooking crucial information. To address these issues, future research can explore the implementation of joint training methods that incorporate both rationale loss and answer loss. Moreover, incorporating token rank information, as annotated within the reasoning chains of the training data, into the training process may improve the performance of fine-tuning CoT in language models.

\subsection{Reflections on the Reasoning Chains from the Dataset}
The reasoning chains in the training data play a crucial role in CoT fine-tuning. There exists a trade-off between templated reasoning chains and those of diverse forms. The use of templated forms allows for models to learn the reasoning skeleton more efficiently, enabling them to focus more on learning the knowledge and the relationship between the question and the answer. However, this approach results in a lack of diversity in the produced CoT and is therefore not suitable for implementing techniques such as self-consistency  \citep{cot_wei_sc} to further improve performance. On the other hand, utilizing more diverse training data poses a challenge for the model to learn effectively. Additionally, the annotation of reasoning chains for existing benchmarks requires a significant amount of effort. Human annotation is costly and time-consuming, and the reasoning chains annotated by humans may not adhere to the correct CoT format. One potential approach to replace human annotation is to utilize large language models to generate reasoning chains. However, the inference of LLMs is also costly, and their performance is not perfect. The CoT produced by large language models may still be incorrect and misleading, therefore requiring human verification.
\section{Conclusion}
This paper presents a method using fine-tuning to enhance the reasoning abilities of a language model. Additionally, we incorporate a CoT filter that can identify and exclude invalid CoTs to form SelF-Reasoner, thereby preventing invalid CoTs from negatively impacting the final answer prediction. The proposed SelF-Reasoner shows a significant performance improvement compared to two-level pipeline approaches, guaranteeing interpretability while maintaining accuracy. In addition, our approach has been shown generally effective across datasets, which achieves consistent performance gains on the ScienceQA, ECQA, and LastLetter datasets. We also conduct an in-depth analysis of the reasoning chains and training data to identify obstacles to achieving perfect CoT. 

%More templated CoT can relieve the limited capacity of the small language model, while reducing the diversity of generated CoT.
%\clearpage
\section*{Limitations}
% The evaluation of our method has been limited to a single dataset. Thus it is not sufficient to fully evaluate its effectiveness.

Three limitations may be addressed in future studies. The first limitation is the increased model size over a single model because we need to train an additional filter. The second limitation is that the filter may still spare incorrect rationales, which can be improved by more effective negative sampling strategies in the filter training. The third limitation is that we evaluate our method in three datasets because most datasets lack effective annotated CoT for fine-tuning. 

% Improvements to the CoT filter are necessary. Furthermore, we have not yet conducted experiments on larger models to illustrate the scaling law. Additionally, the use of an ensemble model increases the number of parameters by a factor of three compared to that of a single model. The effectiveness of the reasoning chain used in our method remains uncertain.
% \section*{Acknowledgement}
% % ?
 \section*{Ethical Considerations}
The primary ethical concern related to this work is the potential for bias in the content generated by the fine-tuned language models used. However, because the focus of this work is the science or commonsense question-answering task, which relies on objective world knowledge and facts rather than subjective statements, the issue of bias is not as significant.

\section*{References}
\vspace{-0.40in}
\bibliography{anthology,custom}
\bibliographystyle{lrec-coling2024-natbib}

 \clearpage
 
 \appendix
 \section{Experiment Details}
\subsection{Dataset}
\label{Appendix: Dataset}
We provide more details on the datasets used in our experiments.

ScienceQA \citep{lu2022learn} is a multimodal dataset annotated with reasoning chains by human experts. It is available at \href{https://scienceqa.github.io/}{https://scienceqa.github.io/}. On ScienceQA, a data example consists of multimodal question-answering information annotated with background lecture and explanation. We use the captions provided in the dataset, which are generated by the model based on ViT \citep{vit} and GPT-2~\citep{gpt2} to replace the visual context following \citep{lu2022fantastically}. The task is formulated as a text-to-text problem where the input $Q$ is a concatenation of question, context, and options and the output is the reasoning chain $R$ or answer $A$ depending on the role of the module as defined in Section~\ref{sec:approach}. %More demonstrations are shown in Figure~\ref{tab:ScienceQA}

ECQA is a human-annotated version of CommonsenseQA~\citep{talmor-etal-2019-commonsenseqa}. The original CommonsenseQA is available at \href{https://www.tau-nlp.sites.tau.ac.il/commonsenseqa}{https://www.tau-nlp.sites.tau.ac.il/commonsenseqa}. ECQA version is available at \href{https://github.com/dair-iitd/ECQA-Dataset}{https://github.com/dair-iitd/ECQA-Dataset}. PINTO~\citep{wang2023pinto} also provides a version annotated by LLM. %Examples with predicted rationale from ECQA are shown in Table~\ref{tab:ECQA}.

LastLetter is a task to concatenate the last letter of the given words. We constructed the dataset from the Google-10000-English repository. We sample 10,000 groups for training and 5,000 for tests (averagely split ranging from 1 to 5 words). The test words are not seen in training words.

We provide the dataset statistics in Table~\ref{tab:statistics}.
\begin{table}[ht]
    \centering
    % mixing matrix
    % \vspace{-5mm}
    % \setlength{\tabcolsep}{20pt}
    % \resizebox{\textwidth}{!}%
    \small
    \setlength{\tabcolsep}{2pt} 
    {
    \centering
    \caption{Dataset statistics used in our experiments.}
    \label{tab:statistics}
        % \resizebox{\columnwidth}{!}
        {
    \begin{tabular}{lccc}
    \toprule
      Dataset & Train &  Validation &  Test \\
     \midrule
      ScienceQA & 12726 & 4241 & 4241 \\
      ECQA & 8520 & 1221 & 1221 \\
      LastLetter & 10000 & 5000 & 5000 \\
         \bottomrule
    \end{tabular}}}
    \vspace{-0.20in}
\end{table}

\subsection{Implementation Details}
\label{Appendix: Implementation Details}
We fine-tune the UnifiedQA for 20 epochs with a learning rate of 4e-5 and a weight decay of 0.01. We use batches of 4. We trained the base and large SelF-Reasoners on Nvidia RTX 2080Ti for 20 and 60 hours, respectively. The reported results are from models trained for 20 epochs. The maximum input sequence length is 512. 
\subsection{Verifier Training Details}
\label{Appendix: Verifier Training Details}
%We use the T5 \citep{t5} encoder extended with a linear classification layer as our verifier model. We load the convergent checkpoint of the answerer as the initial parameters before fine-tuning if possible. It can accelerate the training procedure and improve convergent accuracy.

To construct the training data of the verifier, we run the pipeline model on the training data. The instances with reasoning chain leading to correct predictions will be labeled as positive and those leading to incorrect predictions will be labeled as negative. To balance the number of positive and negative instances, we use a subpar reasoner which scores less precise in the training set to generate invalid instances.\footnote{In ScienceQA, the convergent two-level pipeline can reach a score of 90\% in the training set. So we use a subpar reasoner which scores 60\% in the training set to generate invalid instances.}

\subsection{Discussion on Different Vanilla Fine-tuning Result Compared to \citet{lu2022learn}}
\label{Appendix: Gap}
The difference in the training setting can contribute to the discrepancy. Although we use the same training batch size and learning rate as \citet{lu2022learn}, there are still variances in the details. We have a rating decay of 0.01, and we predict the answer directly instead of the choice. Furthermore, the difference in training time and convergence state also influences performance. Our result is double-checked by rerunning the experiments.

\subsection{Confusion Matrice}
\begin{table}[ht]
    \centering
    % mixing matrix
    % \vspace{-5mm}
    % \setlength{\tabcolsep}{20pt}
    % \resizebox{\textwidth}{!}%
    \small
    \setlength{\tabcolsep}{2pt} 
    {
    \centering
        % \resizebox{\columnwidth}{!}
        {
    \begin{tabular}{lccc}
    \toprule
      Reasoner & Method Used &  Correct Otherwise &  Both Fail \\% & sum\\
     \midrule
      \multirow{2}{4em}{Base} & Directly Predict &154 & 358 \\%& 512 \\
       & Extract & 71 & 119 \\% & 190  \\
     \midrule 
      \multirow{2}{4em}{Large} & Directly Predict &124 & 277 \\%& 401 \\
       & Extract& 38 & 102 \\% & 140  \\
         \bottomrule
    \end{tabular}}}
    \caption{Confusion matrices of the SelF-Reasoner on the incorrect cases. Method Used refers to the adopted method to predict the answer. Correct Otherwise means if the other method were used, the question could be solved. Both Fail means neither method can solve the question.}
    \label{tab:mixing}
    \vspace{-0.20in}
\end{table}

\subsection{Upper Bound of the SelF-Reasoner}
\label{Appendix: upperbound}
The confusion matrices in Table~\ref{tab:mixing} demonstrate the errors made by SelF-Reasoner. There is a certain amount of complex questions that can not be solved using both methods, indicating that the upper bound of SelF-Reasoner with an ideal filter is approximately 89\% for base and 91\% for large.

\subsection{Incorrect CoT Format}
\label{Appendix: CoTFormat}
One typical example is sorting words in alphabetical order, as seen in Figure~\ref{fig:demo_alpha}. The background lecture about alphabetical order is too extensive for the model to process, and the solution part is overly simplistic, resembling a fill-in-the-blank task rather than a logical step-by-step problem-solving process. An example of an expected reasoning chain is also provided in Figure~\ref{fig:demo_alpha}. Additionally, we find that some solutions present the answer first, followed by the explanation.
\definecolor{myblue}{rgb}{0.01, 0.5, 0.73}
\definecolor{myorange}{rgb}{0.85, 0.40, 0.13}
\begin{figure}[ht]
\centering
\includegraphics[width=0.9\linewidth]{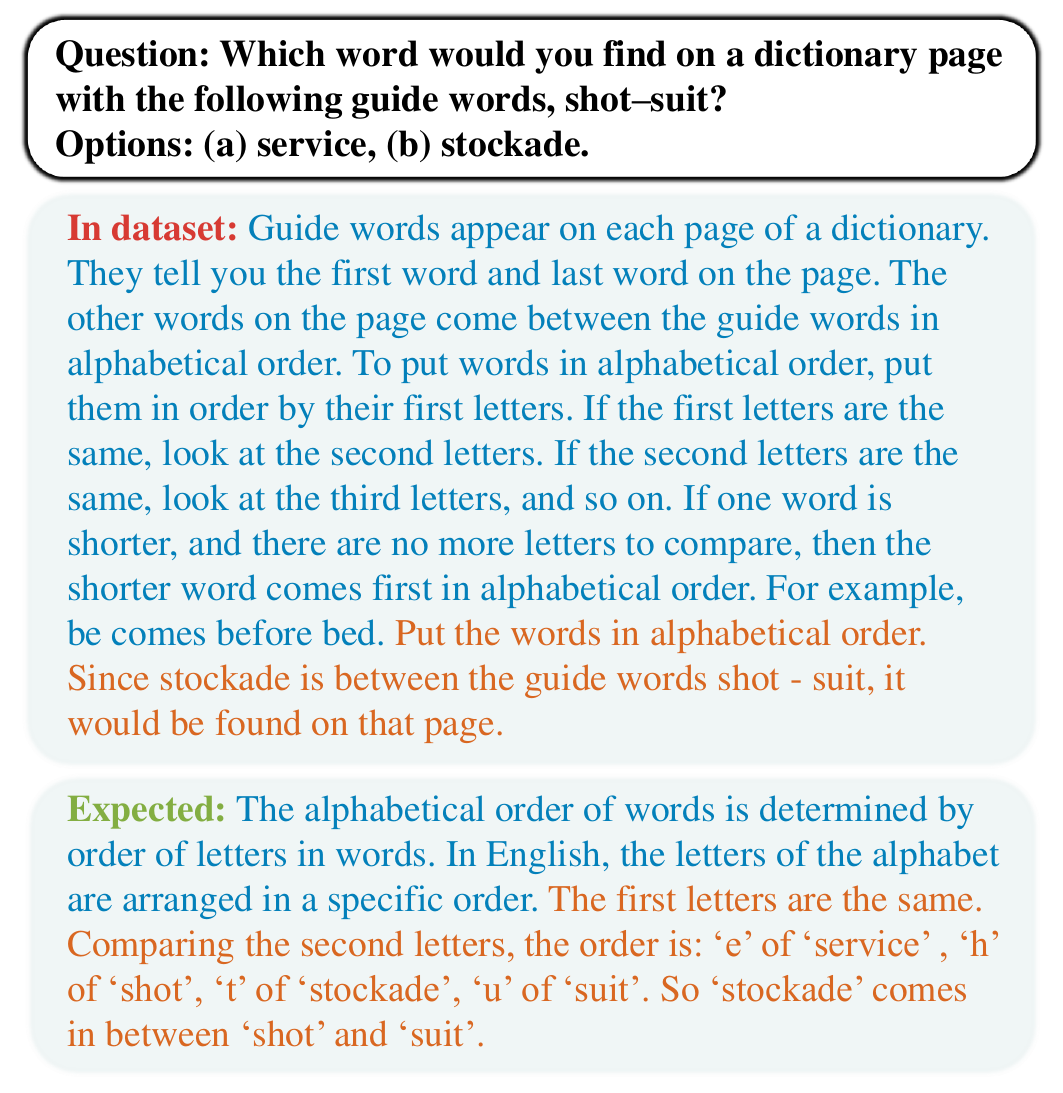}
\caption{An example that does not conform to the CoT format and an expected reasoning chain. \textcolor{myblue}{Blue texts} are the knowledge lectures and \textcolor{myorange}{Orange texts} are the solutions.}
\label{fig:demo_alpha}
\end{figure}
\subsection{Human Evaluation}
\label{Appendix: HumanEval}
In the human evaluation process, the specifications of the three metrics are as follows. \textbf{Complete} means the model generates a complete CoT (No early stop or repeating). \textbf{Entailment} means the extracted answer entails the CoT. \textbf{Correctness} means the reasoning chain is correct and relevant to the question (No irrelevant entities or verbs). 50 examples from each data split are sampled randomly for human evaluation.

\subsection{Case Study}
\label{Appendix:case}
Following is a case from ECQA. The model generates a half-correct CoT. However, the answerer is misled by the generated CoT and gives an incorrect prediction. However, in SelF-Reasoner, the generated CoT is filtered and the answerer gives a correct prediction based on the question.

\begin{quote}
question: Where can you put a picture frame when it's not hung vertically?
choices: (A) wall. (B) newspaper (C) table (D) Car

generated CoT: \textbf{Picture frames are hung vertically on the wall.(Correct part)} Picture frames are not hung vertically on the art show. \textbf{Newspaper or table is not a place where a picture frame can be put.(Misleading part)} Car is not a place where a picture frame can be put.

ground truth CoT: Pictures frames are usually either hung on a wall or kept on a flat surface. A table has a flat surface. Art show usually does not have pictures that are not hung on the wall. Wall is a vertical surface. It is not possible to avoid hanging a picture vertically on a wall. Newspaper is an irrelevant option. Pictures are not generally hung in a car.

prediction: Answer: The answer is wall.
reference: Answer: The answer is table.
\end{quote}

More cases are shown in Table~\ref{tab:ScienceQA}, Table~\ref{tab:ECQA}, and Table~\ref{tab:demo_lastletter}.
\begingroup
\begin{table*}[htb]
    \centering\small
    \caption{
    Demonstrations from ScienceQA. 
    }
    \label{tab:ScienceQA}
    \begin{tabular}{p{0.96\linewidth}}
        \toprule
        
\textbf{Q:} Which of these organisms contains matter that was once part of the lichen? Choices: (A) bilberry (B) mushroom. \\
\vspace{-2mm}
\textbf{Predicted rationale:} Use the arrows to follow how matter moves through this food web. For each answer choice, try to find a path of arrows that starts from the lichen. There are two arrows pointing to the mushroom. One arrow starts from the bear sedge, and the other arrow starts from the lichen. Neither the bear sedge nor the lichen has an arrow pointing to it. So, in this food web, matter does not move from the lichen to the mushroom. \\
\vspace{-2mm}
\textbf{Ground truth rationale:}  Use the arrows to follow how matter moves through this food web. For each answer choice, try to find a path of arrows that starts from the lichen. No arrow points to the bilberry. So, in this food web, matter does not move from the lichen to the bilberry.\\
\vspace{-2mm}
\textbf{A:} The answer is (B) mushroom. \\
\vspace{3mm}

\textbf{Q:} Think about the magnetic force between the magnets in each pair. Which of the following statements is true? Choices: (A) The magnitude of the magnetic force is the same in both pairs. (B) The magnitude of the magnetic force is smaller in Pair 1. (C) The magnitude of the magnetic force is smaller in Pair 2. \\
\vspace{-2mm}
\textbf{Predicted rationale:} The magnets in Pair 2 attract. The magnets in Pair 1 repel. But whether the magnets attract or repel affects only the direction of the magnetic force. It does not affect the magnitude of the magnetic force. Magnet sizes affect the magnitude of the magnetic force. Imagine magnets that are the same shape and made of the same material. The smaller the magnets, the smaller the magnitude of the magnetic force between them. Magnet A is the same size in both pairs. But Magnet B is smaller in Pair 2 than in Pair 1. So, the magnitude of the magnetic force is smaller in Pair 2 than in Pair 1.  \\
\vspace{-2mm}
\textbf{Ground truth rationale:}  Magnet sizes affect the magnitude of the magnetic force. Imagine magnets that are the same shape and made of the same material. The smaller the magnets, the smaller the magnitude of the magnetic force between them. Magnet A is the same size in both pairs. But Magnet B is smaller in Pair 2 than in Pair 1. So, the magnitude of the magnetic force is smaller in Pair 2 than in Pair 1.\\
\vspace{-2mm}
\textbf{A:} The answer is (C) The magnitude of the magnetic force is smaller in Pair 2. \\
\vspace{3mm}

\textbf{Q:} Which of the following contains a vague pronoun reference? Choices: (A) Abby was surprised when the article said that at least one-third of American adults get less than seven hours of sleep each night. (B) Abby was surprised when it said that at least one-third of American adults get less than seven hours of sleep each night.\\
\vspace{-2mm}
\textbf{Predicted rationale:}  The second answer choice contains a vague pronoun reference. The pronoun it could refer to the article or the night. The first answer choice shows a possible correction for the vague pronoun reference. It has been replaced with the article. Abby was surprised when the article said that at least one-third of American adults get less than seven hours of sleep each night. \\
\vspace{-2mm}
\textbf{Ground truth rationale:}  The first answer choice contains a vague pronoun reference. The pronoun it is used without its antecedent. The second answer choice shows a possible correction for the vague pronoun reference. It has been replaced with the article. Abby was surprised when the article said that at least one-third of American adults get less than seven hours of sleep each night.\\
\vspace{-2mm}
\textbf{A:} The answer is (B) Abby was surprised when it said that at least one-third of American adults get less than seven hours of sleep each night.\\
\bottomrule
    \end{tabular}
\end{table*}
\endgroup

\begingroup
\begin{table*}[htb]
    \centering\small
    \caption{
    Demonstrations from ECQA. 
    }
    \label{tab:ECQA}
    \begin{tabular}{p{0.96\linewidth}}
        \toprule
        
\textbf{Q:} Reading newspaper one of many ways to practice your what? Choices: (A) literacy (B) knowing how to read (C) money (D) buying (E) money bank. \\
\vspace{-2mm}
\textbf{Predicted rationale:} Reading newspaper is one of many ways to practice your literacy. Knowing how to read is not practiced. Money is not practiced.Buying is not practiced. Money bank is not a way to practice. \\
\vspace{-2mm}
\textbf{Ground truth rationale:}  Literacy means competency. Reading newspaper provides daily new information to oneself. One can maintain literacy with reading different information daily from newspaper. One cannot read newspaper without knowing how to read. Reading newspaper does not help in achieving other options.\\
\vspace{-2mm}
\textbf{A:} The answer is (A) literacy. \\
\vspace{3mm}

\textbf{Q:} If you are prone to postpone work what will you have to do in order to finish on time? Choices: (A) eat (B) hasten (C) antedate (D) bring forward (E) advance. \\
\vspace{-2mm}
\textbf{Predicted rationale:} If you are prone to postpone work, you will have to have to have a hasten in order to finish on time. Eat is not true as it is not necessary to finish on time. Antidate is not true as it is not necessary to finish on time. Bring forward is not true as it is not necessary to finish on time. Advance is not true as it is not necessary to finish on time. \\
\vspace{-2mm}
\textbf{Ground truth rationale:}  If you have to complete the work on time then you have to speed-up. Antedating will do work early than required, but not on time. Eating, bringing forward and working in advance are also not helpful to do work on time.\\
\vspace{-2mm}
\textbf{A:} The answer is (B) hasten. \\
\vspace{3mm}

\textbf{Q:} Aside from water and nourishment what does your dog need? Choices: (A) bone (B) charm (C) petted (D) lots of attention (E) walked. \\
\vspace{-2mm}
\textbf{Predicted rationale:} Bone is a solid surface of the body that is needed for growth, development, and development. Bone is a part of the body and is not needed for growth, development, or development. Bone is the part of the body and not the body part. Charm is a charm and not the body part. Petted is a human pet and not a body part. Lots of attention is the amount of time and attention given to a dog and not the body part. Walk is to walk and not the body part. \\
\vspace{-2mm}
\textbf{Ground truth rationale:}  Lots of attention is a special care or notice taken of someone or something, the regarding of someone or something as interesting or important. Aside from water and nourishment, our dog needs lots of attention. Bone is a treat that our dog will enjoy, and bone is nourishment and not what our dog needs the most. Charm is a quality of fascinating others and our dog doesn\u2019t have to fascinate others. Petted is to show affection and not a care that a dog requires. Walked is moved at a regular pace and that a dog can do by his own and not what he needs from you.\\
\vspace{-2mm}
\textbf{A:} The answer is (D) lots of attention. \\
\vspace{3mm}

\textbf{Q:} Where can you put a picture frame when it's not hung vertically? Choices: (A) art show (B) wall (C) newspaper (D) car (E) table. \\
\vspace{-2mm}
\textbf{Predicted rationale:}  Picture frames are hung vertically on the wall. Picture frames are not hung vertically on the art show. Newspaper or table is not a place where a picture frame can be put. Car is not a place where a picture frame can be put. \\
\vspace{-2mm}
\textbf{Ground truth rationale:}  Pictures frames are usually either hung on a wall or kept on a flat surface. A table has a flat surface. Art show usually does not have pictures that are not hung on the wall. Wall is a vertical surface. It is not possible to avoid hanging a picture vertically on a wall. Newspaper is an irrelevant option. Pictures are not generally hung in a car.\\
\vspace{-2mm}
\textbf{A:} The answer is (E) table. \\
\bottomrule
    \end{tabular}
\end{table*}
\endgroup

\begingroup
\begin{table*}[htb]
    \centering\small
    \caption{
    Demonstrations from LastLetter Task. 
    }
    \label{tab:demo_lastletter}
    \begin{tabular}{p{0.96\linewidth}}
        \toprule
        
\textbf{Q:} immigrants, editorials, hierarchy \\
\vspace{-2mm}
\textbf{Predicted rationale:} The last letter of the first word 'immigrants' is's'. The last letter of the second word 'editorials' is 's'. The last letter of the third word 'hierarchy' is 'y'. \\
\vspace{-2mm}
\textbf{Extracted Answer:} The answer is ssy. (Correct) \\
\textbf{Directly predicted Answer:} The answer is ssy. (Correct) \\
\textbf{Ground Truth Answer:} The answer is ssy.\\
\vspace{3mm}

\textbf{Q:} poison, dame, cornell \\
\vspace{-2mm}
\textbf{Predicted rationale:} The last letter of the first word 'poison' is 'n'. The last letter of the second word 'dame' is 'e'. The last letter of the third word 'cornell' is 'l'. \\
\vspace{-2mm}
\textbf{Extracted Answer:} The answer is nel. (Correct)\\
\textbf{Directly predicted Answer:} The answer is nell. (Inorrect) \\
\textbf{Ground Truth Answer:} The answer is nel.\\
\vspace{3mm}

\textbf{Q:} departmental, dome, pressed, fascinating \\
\vspace{-2mm}
\textbf{Predicted rationale:} The last letter of the first word 'departmental' is 'l'. The last letter of the second word 'dome' is 'e'. The last letter of the third word 'pressed' is 'd'. The last letter of the forth word \textcolor{red}{'funny' is 'y'} .\\
\vspace{-2mm}
\textbf{Extracted Answer:} The answer is ledy. (Incorrect, the rationale part is wrong) \\
\textbf{Directly predicted Answer:} The answer is ledg. (Correct) \\
\textbf{Ground Truth Answer:} The answer is ledg.\\
\vspace{3mm}

\textbf{Q:} bahrain, joyce, calibration, ambien \\
\vspace{-2mm}
\textbf{Predicted rationale:} The last letter of the first word 'bahrain' is 'n'. The last letter of the second word 'joyce' is 'e'. The last letter of the third word 'calendar' is 'r'. The last letter of the forth word 'ambien' is 'n'. \\
\vspace{-2mm}
\textbf{Extracted Answer:} The answer is \textcolor{red}{nern}. (Incorrect, extraction part is wrong) \\
\textbf{Directly predicted Answer:} The answer is nenn. (Inorrect) \\
\textbf{Ground Truth Answer:} The answer is ssy.\\
\bottomrule
    \end{tabular}
\end{table*}
\endgroup

\end{document}